# RESOLVING PART-OF-SPEECH AMBIGUITY IN THE GREEK LANGUAGE USING LEARNING TECHNIQUES

**Georgios Petasis, Georgios Paliouras, Vangelis Karkaletsis, Constantine D. Spyropoulos and Ion Androutsopoulos**

**Software and Knowledge Engineering Laboratory**
**Institute of Informatics and Telecommunications, N.C.S.R. "Demokritos",**
Tel: +301-6503197, Fax: +301-6532175
E-mail: {petasis, paliourg, vangelis, costass, ionandr}@iit.demokritos.gr

## ABSTRACT

This article investigates the use of Transformation-Based Error-Driven learning for resolving part-of-speech ambiguity in the Greek language. The aim is not only to study the performance, but also to examine its dependence on different thematic domains. Results are presented here for two different test cases: a corpus on "management succession events" and a general-theme corpus. The two experiments show that the performance of this method does not depend on the thematic domain of the corpus, and its accuracy for the Greek language is around 95%.

## INTRODUCTION

The aim of the work presented in this paper is to study the performance of Transformation-Based Error Driven (TBED) learning for resolving Part-of-speech (POS) ambiguity in the Greek language, as well as to examine its dependence on the thematic domain. Part-of-speech taggers are classifiers that aim to assign unambiguous tags to words in electronic documents, according to the part of speech in which they belong, i.e., verbs, nouns, adjectives etc. Part-of-speech tagging is a practical application with many uses, such as in information extraction, machine translation and speech recognition. As a result, a large number of different approaches have been applied to POS tagging, such as Markov models (Weischedel, 1993), decision trees (Black, 1992), (Cardie, 1994), (Daelemans, 1996), (Orphanos and Christodoulakis, 1999) connectionist machines (Schmid, 1994), nearest-neighborhood algorithms (Daelemans, 1996), TBED (Brill, 1995) and maximum-entropy techniques (Ratnaparkhi, 1996). All these methods seem to achieve roughly comparable accuracy in the context of English language. This accuracy is usually in the range of 94-98%.

For our experiments we used a publicly available[1] POS tagger implementation from the author of the TBED learning technique: the Brill tagger (Brill, 1995). We have chosen to use the TBED learning technique and the Brill tagger mainly for its practical performance: Brill tagger has shown good results for English and there is evidence that rule-based taggers can achieve better results than stochastic ones in the English language (Samuelsson and Voutilainen, 1997). Additionally, there are few successful attempts to train the Brill tagger to languages other than English, such as German (Schneider and Volk, 1998), French (Chanod and Tapanainen, 1995), Italian (Basili et al., 1996) and Estonian (Schneider, 1997).

According to the TBED part-of-speech tagging technique, an initial tag is assigned to each word. This tag is the most likely (frequent) tag for the word if the word is known. Frequency information is stored in a *lexicon*, which has been constructed during the training phase and contains all the words in the training corpus associated with their most frequent POS tag, as was measured from the training corpus. In the case of an unknown word, a default rule is used initially for tagging. For the English language, this default initial tagging rule is the following:

    **IF** *(word starts with a capital letter)* **THEN**
        classify word as a singular proper noun
    **ELSE**
        classify word as a singular noun.

---

[1] The Brill tagger is available from its author at http://www.cs.jhu.edu/~brill.

Once the assignment of initial POS tags has been completed for all the words in the corpus, an ordered sequence of *lexical rules* is applied to the corpus. Each one of these lexical rules operates only on a single word, and its preconditions *consider only morphological cues*. For example, a typical lexical rule has the following form:

   **IF** *(word ends in "ed")* **THEN**
       classify word as a verb in the past tense.

After the application of lexical rules has been completed, an ordered list of *contextual rules* is applied to the corpus. Each of these rules can change the tag assigned to a word according to the context in which the word appears. The environment used for changing a word tag consists of the words and tags within a window of four words, including the word under examination.

All of the resources needed (the lexicon, the lexical and the contextual rule set) are created during the training phase of the Brill tagger. The Brill tagger involves two training stages. In the first stage, rules are learned from the training corpus to assign POS tags. These rules operate on word types. The tag chosen for each word holds for all occurrences of the word in the corpus. The output of this phase is a lexicon, containing every word in the training corpus associated with its most frequent tag, and an ordered list of lexical transformation rules that are based on morphological information. The search space examined by the Brill tagger and as a result the form and complexity of the produced lexical rules, is defined by *a lexical rule template*, which describes all the possible forms of the rules that can be produced. In the second training phase, rules are learned to use contextual cues to improve tagging accuracy. An example of such a rule is the following:

   **IF** *(current word tagged as a verb* **AND** *previous word tagged as a determiner)* **THEN**
       tag current word as noun.

These rules operate on individual word tokens. The output of this phase is also an ordered list of transformation rules, that are based on contextual information such as the current tags of the surrounding words or the surrounding words themselves, within a window of three words. A *contextual rule template* also describes all possible contextual rules that can be derived in this training phase.

The work presented in this paper has been performed in the context of the research project GIE (Greek Information Extraction), a bilateral project between NCSR "Demokritos" and University of Sheffield, funded by the Greek General Secretariat of Research &Technology and the British Council.

Section 2 presents some topics of interest when applying the Brill tagger in the Greek language. Sections 3 and 4 present the experimental results of the application of Brill tagger to two Greek corpora: a general-theme Greek corpus and a Greek corpus on "management succession events". Finally, section 5 presents some conclusions about the usability of this learning method in disambiguation of POS ambiguity in the context of the Greek language.

## EXPERIMENTAL SETTING

In order to examine the behavior of the Brill tagger in the Greek language, a new tag set had to be specified for the Greek language (Karkaletsis et al., 1998). Because the Greek language is a highly inflectional language, a compromise had to be made, regarding the features of the language that should be described by the new tagset i.e., cases for nouns, adjectives and verbs, mood for verbs, etc. We finally decided to use a rather limited tag set for the Greek language, containing only 58 tags for efficiency reasons. The original tag set that is used by the Brill tagger for the English language contains 48 tags.

Due to our interest in examing the learning procedure of the Brill tagger under different thematic domains, in this experiment we used two totally different corpora, one of which is domain specific while the other is not. The first one contained news articles from the Greek newspaper *"Advertising Week" ("Διαφημιστική Εβδομάδα", http://www.adweek.gr)* and its domain was "management succession events". The corpus contains texts on personnel leaving or joining companies for the period from 1/96 until 12/98. The corpus size was about 65,000 words. Part of this corpus (about 36.000 words) was hand-tagged in order to be used for this experiment. The second text corpus is a general-theme hand-tagged corpus, which was provided by the WCL[2] Laboratory of Patras University. The size of this corpus is about 125.000 words.

A particularity of the corpus on "management succession events" is the existence of a large number of foreign (mainly English) words, such as organizations and person names. This characteristic has affected the choice of the default initial tagging rule for the Brill tagger. For the English language, the default initial tagging rule is typically the following:

---

[2] Wired Communications Laboratory, Dept. of Electrical and Computer Engineering, University of Patras, Greece.

    **IF** *(word starts with capital letter)* **THEN**
        classify word as a singular proper noun
    **ELSE**
        classify word as a singular noun.

For the Greek language, this rule was changed to:

    **IF** *(word starts with an English character)* **THEN**
        classify word as a foreign word
    **ELSEIF** *(word starts with a capital Greek letter)* **THEN**
        classify word as a singular male proper noun
    **ELSE**
        classify word as a singular female noun.

The same initial tagging rule was applied to both experiments that are presented in the rest of the paper.

The performance of the tagger is always measured on unseen data. In order to derive a robust and unbiased estimate of the method's performance, we used 10-fold cross validation at each individual experiment. According to this evaluation method, the corpus is split into ten, equally sized sub-corpora and the final result is the average over ten runs. In each run, nine of the ten sub-corpora are used to train the tagger and the tenth is held out for the evaluation. Thus, each accuracy figure presented in the following sections is the average over ten runs, rather than a single train-and-test result, which can often be accidentally high or low.

## PART-OF-SPEECH TAGGING IN THE GENERAL-THEME CORPUS

In the first test case, we used part of the general-theme corpus, provided by the WCL Laboratory of Patras University. The size of the complete corpus is about 125.000 words, and is organized in a single file, where each line corresponds to a single sentence. Each word of this corpus has been tagged using an extremely rich, full-featured tagset for the Greek language. A new version of the original corpus was created, where each tag was mapped onto our limited tagset. Then, the sentences of the newly created corpus were shuffled using a randomizer. The reason for doing so is the structure of the corpus, which is composed of small sentence groups that belong to the same thematic domain.

In this test case, we evaluated the Brill tagger using 10-fold cross validation over different corpus sizes. The results are shown in Figure 1. The error bars correspond to the standard deviation of the average accuracy over the ten runs of the 10-fold cross validation. As expected, tagging accuracy increases as the corpus size increases. Accuracy seems to stabilize around 95%, which is lower than the highest reported accuracy for English. This is mainly due to the tagging difficulties for the Greek language, such as morphological complexity. The Brill tagger seems to perform well when the corpus size used for training is greater than 18.000 words.

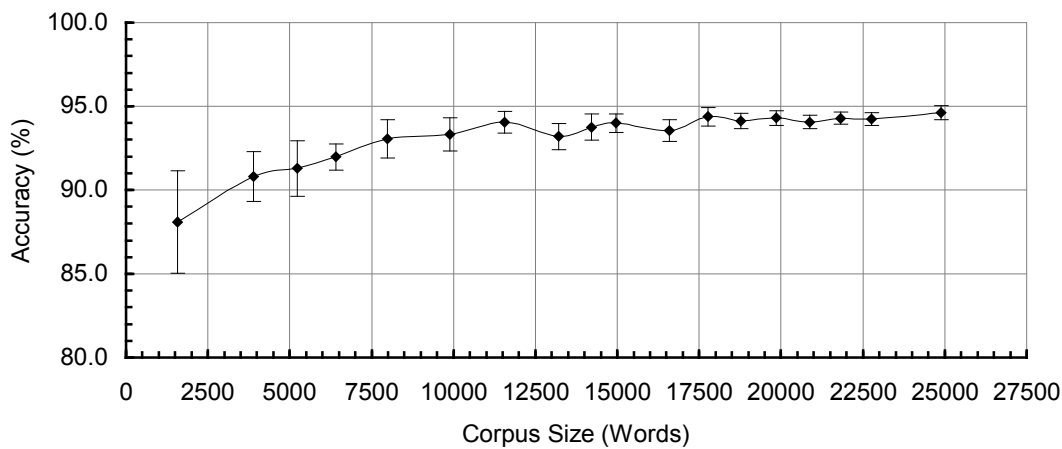

**Figure 1:** Brill tagger accuracy versus corpus size (general theme corpus).

Another point of interest, is the examination of the number of learned rules. Usually, high performance of a learning task when combined with a small rule set size indicates robustness of the learning task. Large numbers of learned rules often indicate problems in the learning task that the algorithm tries to overcome through overfitting of the training data. In other words, a large number of learned rules is usually a sign that the learning algorithm tries to "remember" the training data, instead of discovering the underlying assumptions that govern it. As the task of training the Brill tagger involves two different learning sub-tasks (learning lexical rules and learning contextual rules), we had the opportunity to examine them separately. As was explained previously, the lexical rules correspond to the morphology and the contextual rules to the grammatical and syntactic features of the language. Thus, by examing the sizes of lexical and contextual rule sets separately, we can isolate any potential problems to the morphologic or the grammatical/syntactic properties of the language.

The size of the learned rule set against corpus size is shown in Figure 2. As can be seen from Figure 2, the number of both types of rule (lexical and contextual) increases almost linearly with the corpus size. The number of the lexical rules is much larger than the number of the contextual rules. This large number of lexical rules indicates that the Brill tagger faces difficulties in coping with the morphology of the Greek language. It seems that it is easier for the Brill tagger to derive rules based on grammatical and syntactic properties (contextual rules) than rules based on morphology (lexical rules) in the context of the Greek language. One justification for this phenomenon is that the template used for deriving the lexical rules (lexical rule template) is too restricted for the Greek language.

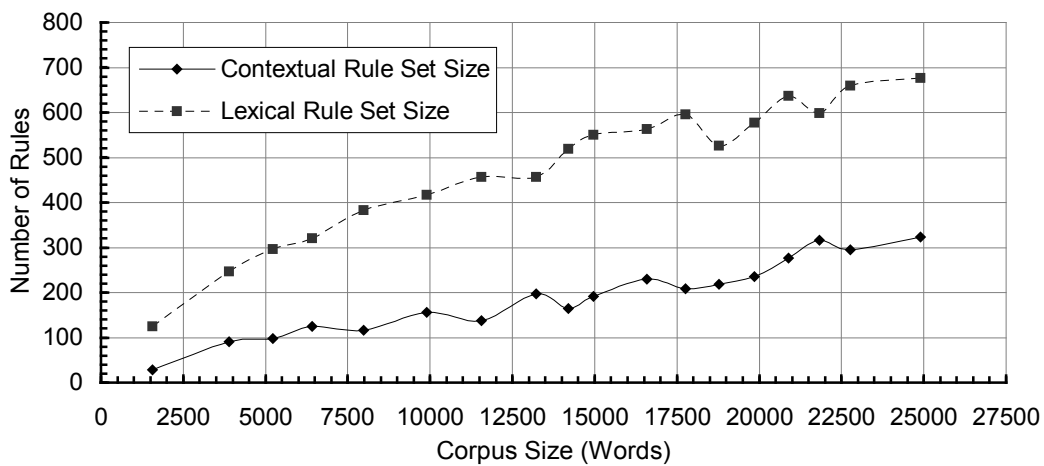

**Figure 2:** Number of Lexical and Contextual Rules versus corpus size (general theme corpus).

The above experiments show that the accuracy of the Brill tagger for the Greek language is around 95%, when applied on a general-theme corpus. The number of the learned rules rises almost linearly with the corpus size. The number of lexical rules is always at a higher level than the corresponding level of the number of learned contextual rules. This fact indicates that the Brill tagger faces a number of difficulties regarding the morphology of the Greek language.

## PART-OF-SPEECH TAGGING IN "MANAGEMENT SUCCESSION EVENTS" CORPUS

In the second test case, we applied the Brill tagger on a corpus from a restricted thematic domain. For this purpose, we used part of the corpus on "management succession events", from the "Advertising Week". The original size of the corpus is about 65.000 words. About 36.000 words were hand-tagged using the same tagset as in the previous test. We evaluated the Brill tagger using 10-fold cross validation over different corpus sizes. The results are shown in Figure 3. As was the case in the previous experiment, tagging accuracy increases as the corpus size increases. Accuracy also seems to stabilize around 95%, although the thematic domain was more restricted and probably the language was more "controlled". As a result, the performance of the Brill tagger in the Greek language does not seem to depend on the thematic domain of the corpus.

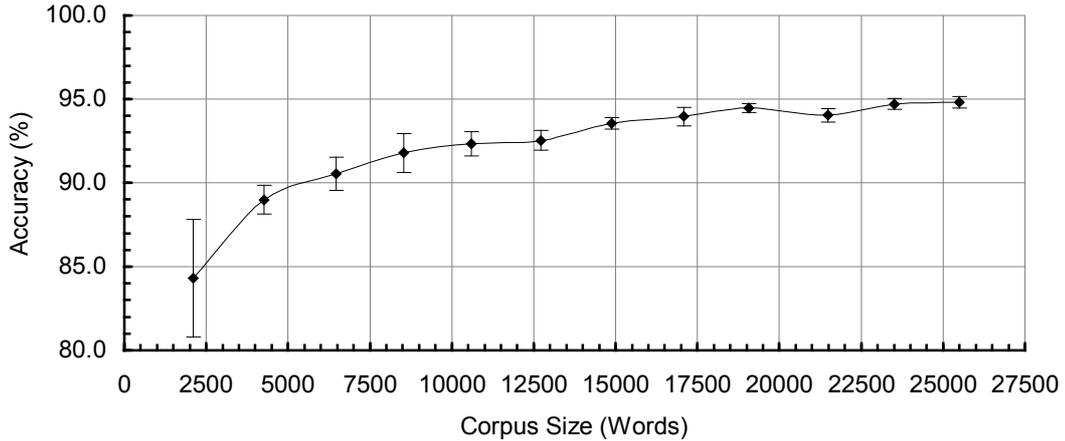

**Figure 3:** Brill tagger accuracy versus corpus size ("management succession events" corpus).

The size of the learned rule sets against the corpus size is shown in Figure 4. As can be seen from Figure 4, the number of both types of rule, lexical and contextual, increases almost linearly with the corpus size, as was also the case in the general-theme corpus. Additionally, as in the general-theme corpus, lexical rules are almost twice as many as contextual rules. Contextual rules are at a similar level as in the general-theme corpus but lexical rules are at lower level than lexical rules in the general-theme corpus. This reduction is to be expected, as the language in this corpus is more restricted and a large number of foreign words exist. All foreign words are assigned the same tag, classifying them as foreign words, and thus are easier to disambiguate.

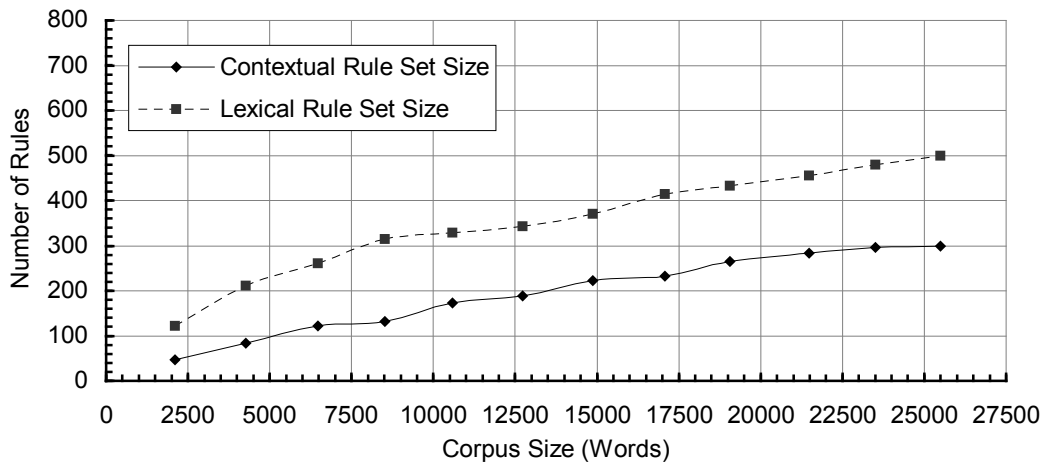

**Figure 4:** Number of Lexical and Contextual Rules versus corpus size ("management succession events" corpus).

The above experiments show that the accuracy of the Brill tagger for the Greek language is around 95%, with minimal dependence on the thematic domain. Our initial expectation that the accuracy of applying the Brill tagger to a corpus with a specific thematic domain would be at higher levels than the accuracy on the general-theme corpus, was not confirmed as this increase has not been observed in our experiments.

## CONCLUSIONS

In the work presented here we have applied a popular machine learning technique, the Transformation-Based Error-Driven learning, to the task of part-of-speech tagging in the context of the Greek language. We have trained the Brill tagger over relatively small-sized annotated Greek corpora and we have found its performance to be around 95%. An important conclusion that can be drawn from our results is that the performance of the Brill tagger does not significantly depend on the domain of the corpus, at least when applied to the Greek language. Thus, porting the tagger to different domains should require minimal effort. Our experiments have shown that the performance of the Brill tagger when applied to the Greek language begins to stabilize when the size of the training corpus approaches to 18.000 words. Any further increase in the size of the training corpus results in a relative small increase in the tagger's performance. Thus, the recommended size of the training corpus is around (or greater than) 18.000 words.

However, the performance of Brill tagger in the Greek language is slightly lower than in the English language. Our main aim for the future is to try to improve the performance of Brill tagger for the Greek language, by trying to isolate the difficulties and solve the problems that arise in the context of the Greek language. Our results provide evidence that the existing lexical rule templates, incorporated in the tagger, are not adequate in coping with the Greek language morphology. In future work we will try to adapt the lexical rule templates to the peculiarities of the Greek language, in order to increase the overall accuracy.